# A Multi-task Two-stream Spatiotemporal Convolutional Neural Network for Convective Storm Nowcasting

*Wei Zhang*
*College of Information Science and Technology Ocean University of China*
*Qingdao, China*
weizhang@ouc.edu.cn

*Houling Liu*
*College of Information Science and Technology Ocean University of China*
*Qingdao, China*
liuhongling@stu.ouc.edu.cn

*Pengfei Li*
*College of Information Science and Technology Ocean University of China*
*Qingdao, China*
lipengfei@stu.ouc.edu.cn

*Lei Han*
*College of Information Science and Technology Ocean University of China*
*Qingdao, China*
hanlei@ouc.edu.cn

**ABSTRACT**

**The goal of convective storm nowcasting is local prediction of severe and imminent convective storms. Here, we consider the convective storm nowcasting problem from the perspective of machine learning. First, we use a pixel-wise sampling method to construct spatiotemporal features for nowcasting, and flexibly adjust the proportions of positive and negative samples in the training set to mitigate class-imbalance issues. Second, we employ a concise two-stream convolutional neural network to extract spatial and temporal cues for nowcasting. This simplifies the network structure, reduces the training time requirement, and improves classification accuracy. The two-stream network used both radar and satellite data. In the resulting two-stream, fused convolutional neural network, some of the parameters are entered into a single-stream convolutional neural network, but it can learn the features of many data. Further, considering the relevance of classification and regression tasks, we develop a multi-task learning strategy that predicts the labels used in such tasks. We integrate two-stream multi-task learning into a single convolutional neural network. Given the compact architecture, this network is more efficient and easier to optimize than existing recurrent neural networks.**

## INTRODUCTION

Severe convective weather includes wind, hail, and heavy rain. Short-term prediction of convective storms in both time and space is desirable [1]. There are currently two main types of nowcasting technologies: extrapolation based on radar data, and numerical weather prediction (NWP) [2]. Given the fundamental lack of understanding of the physical processes of convective weather, NWP cannot accurately predict convective storms. Many extrapolation methods use weather radar to obtain high temporal and spatial resolution convective storm data [3]. However, in mountainous or sparsely populated regions, radar information may be incomplete and many forecasts cannot be verified, restricting the development of radar nowcasting. Today, satellite images are widely used to estimate precipitation, providing valuable nowcasting data[4] [5] [6]. Satellites survey large areas and are minimally affected by the ground environment. However, satellite spatial and temporal resolutions are lower than those of ground- based radars, compromising nowcast accuracy and timeliness. Thus, we derived a network exploiting both types of data to identify strongly convective weather. Inspired by [7], we use three-dimensional (3D) spatiotemporal convolution to extract multivariate spatiotemporal features from radar and satellite data, and pixel-wise data sampling to deal with inconsistencies in spatial resolution. As the classification and regression tasks are correlated, nowcasting accuracy is improved by deriving the learned correlations [8] [9]; the overall generalization error is thus reduced. When performing classification and regression, existing temporal networks require large amounts of data; this increases the number of parameters, which makes it difficult to train the networks. A more concise and effective method is required to accurately predict severe convective storms. Thus, we developed a simple, two-stream multi-task (TSMT) spatiotemporal convolutional neural network (CNN) using pixel-wise data sampling. Regression is used to improve forecasting accuracy and strengthen the network. MT learning improved the predictive accuracy. The contributions of this paper are as follows:

- Pixel-wise sampling: given the low frequency and relatively short duration of convective weather, a typical class-imbalance learning problem is seen. We use pixel- wise sampling to increase the proportion of positive samples in the training set(However, the test set remains its original proportion to be consistent with the real scenarios). This also solves the problems associated with the different spatial resolutions of radar and satellite data.
- A two-stream (TS) network: the TS network tends to reduce the degree of dependence on a single dataset; the two types of data are mutually complementary, enhancing nowcasting accuracy.
- MT learning: we employ a multi-task (MT) learning strategy to improve classification accuracy and reduce the discrepancy between the ground truth and the predicted label.

  The remainder of the paper is organized as follows. In Section 2, we discuss related reports; Section 3 describes our preparatory work; Section 4 deals with the TSMT- CNN; experimental results are presented in Section 5; and Section 6 details our conclusions and future plans.

RELATED WORK

Machine learning techniques have recently been used to address the challenges associated with short-term forecasting. Reference [10] used artificial neural networks (NNs) to fore- cast localized precipitation. Reference [11] developed a deep neural network (DNN) that combined pre-trained predictive models to model the statistics of a set of weather-related variables. Reference [12] employed a DNN to process massive weather data. Reference [13] used traditional machine learning to re-analyze meteorological data, which facilitated nowcasting. Reference [14] developed the ConvLSTM model using continuous two-dimensional (2D) radar images to generate images for precipitation nowcasting, followed by the similar recurrent neural network (RNN) such as TrajGRU model [15] or predRNN [16], RNN becomes popular to deal with spatiotemporal convective nowcasting problem. However, training

RNN is difficult because back-propagation of the gradient through the network may either explode or vanish at each time step [17] [18] [19]. Although long short-term memory (LSTM) or a gated recurrent unit (GRU) can mitigate the problem, it remains difficult to train complicated RNNs featuring LSTM or GRUs using stacked encoder and decoder layers.

Recently, CNNs have been effectively applied for spatiotemporal learning. To learn data patterns directly (i.e., in the absence of handcrafted feature engineering), [20] fused raw radar observations with NWP variables to construct a CNN, termed 3D-SCN. Reference [21] developed a dynamic CNN that used four consecutive 2D radar reflectivity images to produce radar images for the next 10 min. Reference [7] used 3D CNNs to extract spatiotemporal features. We also employ a 3D spatiotemporal CNN to extract radar and satellite characteristics. In addition, inspired by [22], we incorporate Multi-Task (MT) formulation that automatically learns task correlations. MT learning can improve generalization performance by jointly learning several tasks via shared representation [23] [24] [25]. Experimentally, regression tasks increase classification task accuracy.

## PRELIMINARIES

### A. Formulation of Nowcasting Problem

From a machine learning perspective, problems can be regarded as spatiotemporal sequence forecasting problems. Suppose that P variables are observed in a dynamic system; for example, a video stream contains three RGB channels, i.e., three variables. Each variable can be represented by a 2D M*N lattice point at each moment. Given the dimensions of the variables, the P variables collected at any time can be regarded as a 3D tensor X. In the time dimension, we collect observational data at fixed time intervals to create a serial dataset $\{x_1, x_2, x_3, \cdots, x_t\}$. Here, we express radar data as $\{x_{r1}, x_{r2}, x_{r3}, x_{r4}, x_{r5}\}$ and satellite data as $\{x_{s1}, x_{s2}, x_{s3}\}$.

For convective storm nowcasting, 35 dBZ is the threshold widely used by meteorologists. Rainfall of this intensity correlates well with the eventual development of mature cumulonimbus clouds [26] [27](Roberts and Rutledge 2003; Mecikalski and Bedka 2006). From a classification perspective, if the radar echo from a specific location is 35 dBZ for 30 min [13], the sample for that position is labeled 1 (a convective storm will develop in this cell within 30 min). Otherwise, the sample is labeled 0 (no storm within 30 min). Thus, the classification and regression used are:

$$C_k = argmax_{C_k \in \{0,1\}} P(C_k | X_{r1}, \cdots, X_{rt}, X_{s1}, \cdots, X_{st})$$
$$X_{t+1} = argmax_{X_{t+1}} P(X_{t+1} | X_{r1}, \cdots, X_{rt}, X_{s1}, \cdots, X_{st})$$
(1)

### B. Data description

The study area was the Beijing-Tianjin-Hebei triangle. A 2D radar echo map was generated with a time resolution of about 6 min and a spatial resolution of 0.01*0.01. The satellite data are from the Japanese Himawari-8 satellite. Multi-channel satellite images (including near-infrared [NIR] and infrared [IR] channels) were obtained with a time resolution of 10 min and a spatial resolution of 0.02*0.02.

### C. Pixel-wise sampling

Given the low frequency of convective storms, nowcasting is a form of class-imbalance learning [28], where it is necessary to increase the proportion of positive samples or reduce that of negative samples in the training set. Previous studies [14] [15] extrapolated historical radar images to new short-term images. However, it was difficult to adjust the positive:negative sample ratio; convective storms developed only locally according to the radar images. We used a pixel-wise sampling method for simple oversampling of positive samples or undersampling of negative samples; the method also solves the problem posed by the different spatial resolutions of radar and satellite data. The sampling is discussed below in terms of temporal and spatial feature construction.

### D. Pixel-wise temporal feature construction

The first step when developing the data-processing algorithm was selection of candidate features or predictors for training [26] [27] [29]. Satellite IR cloud-top temperature trends are very useful for predicting the onset of convective storms. Thus, we selected 13 IR and NIR frequency bands from among the satellite data as candidate features, and used the radar echo reflectivity data (R values) as candidate features. As shown in Figure 1, 14 variables served as candidate predictors.

The temporal trend contains information on storm growth/decay and plays an important role in convective nowcasting. We constructed the temporal features of the time trends and used five consecutive radar frames $\{x_{t-4}, x_{t-3}, x_{t-2}, x_{t-1}, x_t\}$ and three consecutive satellite frames $\{x_{t-2}, x_{t-1}, x_t\}$ to obtain sequence information.

### E. Pixel-wise spatial feature construction

Weather phenomena are continuous in both time and space. Thus, each pixel is affected by its neighbors, so spatial features are required. As shown in Figure 2, we chose the area around the red (central) pixel for extraction of 54*54 features predicting the center. Then, we moved several pixels and repeated this procedure. Thus, information obtained around the center is considered to facilitate nowcasting.

After construction of temporal and spatial features, we normalized the values to [-1, 1] by setting $X = \frac{(x - x_{min}) * 2}{x_{max} - x_{min}} - 1$, and generated the final dataset.

## THE MODEL

The network architecture and learning configuration of the TS-MTCN are described in this section. The general architecture, shown in Figure 3, includes the following: 1) a TS module; 2) a multivariate data fusion module; and, 3) an MT learning module. Different NN paths are chosen depending on the input data, and spatial and temporal features are gradually extracted. We perform simultaneous image classification and label regression and describe the architectures of the three modules below.

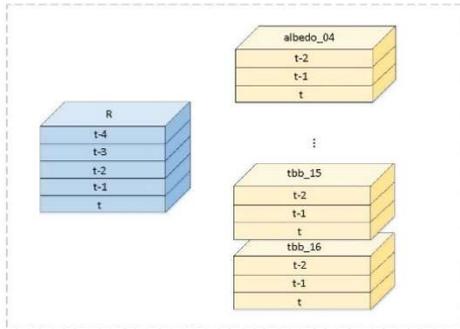

Fig. 1. Radar and satellite input data schematic

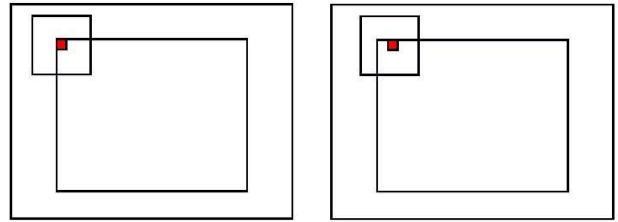

Fig. 2. The red box is a pixel, and the surrounding 54*54 pixels are taken as a sample

## A. The TS network

The TS module is shown in Figure 3. The module accepts two different spatiotemporal data streams. For radar data streams (R values), the input is $I_r = [R_t]_{t=1}^{5}$. The satellite data stream includes 13 observations and the input is $I_s = [P_t]_{t=1}^{3}$, where $P_t$ is defined as:

$$P_t = stack(albedo\_04t, albedo\_05t, \ldots tbb\_16t) \quad (2)$$

To avoid interference between the time and variable axes when extracting features of satellite data, we first extracted the time series features of the variable at an adjacent time via 3D convolution, and then fused the information obtained. We used batch normalization and activation layers to minimize gradient disappearance and accelerate convergence. However, the values for adjacent grids were typically very similar; both $R_t$ and $P_t$ were thus somewhat redundant, but nonetheless compromised the model training. Therefore, we reduced the spatial resolution of $R_t$ and $P_t$ using a pooling layer.

## B. The multivariate data fusion module

Radar data can be very precise but spatial coverage is usually incomplete. Satellites deliver more comprehensive, but less accurate, weather information. The fusion module exploits the advantages of both datasets to improve predictive accuracy. The radar features ($\bar{R}$) and satellite features ($\bar{P}$) are fused to yield:

$$F^{fuse} = stack(\bar{R}, \bar{P})$$
$$\bar{F}^{fuse} = Relu\left(Conv2D(F^{fuse})\right) \quad (3)$$

## C. MT learning

Adaptive average pooling of the data output by the fusion module was performed. Two Fully Connected (FC) modules with fully connected and activation function layers are also included in the model. Finally, a fully connected layer outputs probabilistic predictions using softmax activation. The crossentropy loss is given in Equation 4:

$$L\_c(W, b) = -\frac{\sum_{i=1}^{m}(y_i \log(prob_i) + (1 - y_i)\log(1 - prob_i))}{m} \quad (4)$$

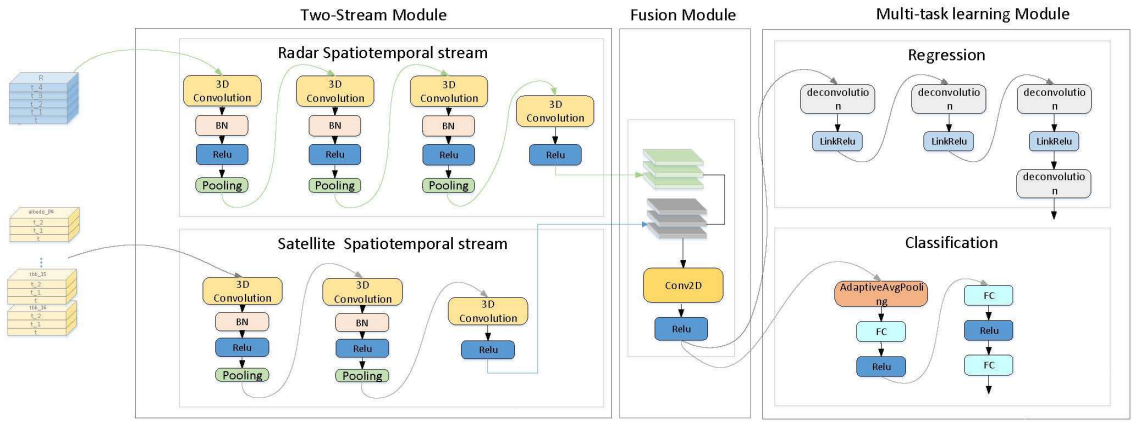

Fig. 3. The overall network architecture of our TS-MTCN

where m is the number of inputs, $y_i$ are the labels, and $prob_i$ is the probability output.

The regression task uses three deconvolution modules (Figure 3) and one deconvolution layer. Given the variable effects of peripheral information, the predicted label must be as close to the ground truth as possible. To achieve this, we reduced the prediction area to 48*48 pixels and introduced a weight matrix (WM). When the radar echo is closer to the median, its influence on median prediction is greater, as is the weight. A WM was created using Equation 5:

$$WM = \begin{bmatrix} 1 & 1 & \cdots & 1 & 1 \\ 1 & 2 & \cdots & 2 & 1 \\ \vdots & \vdots & \ddots & \vdots & \vdots \\ 1 & 2 & \cdots & 2 & 1 \\ 1 & 1 & \cdots & 1 & 1 \end{bmatrix} \tag{5}$$

The initial matrix was normalized employing Equation 6:

$$WM_{ij} = \frac{e^{WM_{ij}}}{\sum_i \sum_j (e^{WM_{ij}})} \tag{6}$$

We used the mean squared error (MSE) method to calculate all pixel losses in the study area, and summed the loss matrix and WM element-wise products to yield the final loss caused by regression, as follows:

$$MSE_{ij} = \left(output_{reg_{ij}} - label_{reg_{ij}}\right)^2$$

$$\text{L\_r}(W, b) = \frac{\sum_{k=1}^{m}(\sum_{i=1}^{n}\sum_{j=0}^{n}(MSE_{ij} * WM_{ij}))}{m} \tag{7}$$

The combined loss equation (Equation 8) was used during training of our MT learning model. α and β vary according to the scenario, and adjust the balance between classification loss and regression loss.

$$F^{fuse}\text{L\_all} = \alpha \text{L\_c} + \beta \text{L\_r} \tag{8}$$

Finally, we present the inference algorithm in Algorithm 1.

---

**Algorithm 1** Multi-tasking weight updates
---
**Input:** radar data: $\{x_{t-4}, x_{t-3}, x_{t-2}, x_{t-1}, x\}$
satellite data: $\{x_{t-2}, x_{t-1}, x_t\}$
**Output:** Model parameters: $W, b$
1: initial $W$ using Gaussian distribution;
2: initial $b = 0$;
3: initial the weight matrix: $WM_{ij} = min\{i + 1, j + 1, 48 - i, 48 - j\}$;
4: Normalize the weight matrix: $WM_{ij} = \frac{e^{WM_{ij}}}{\sum_i \sum_j (e^{WM_{ij}})}$;
5: $count = 0$;
6: **while** ($count < iterations$) **do**
7:     Read a minibatch;
8:     compute classification loss:

$$\text{L\_c}(W,b) = -\frac{\sum_{i=1}^{m}(y_i(\log(prob_i) + (1-y_i)\log(1-pro_{\_i}))}{m};$$

9:     compute regression loss:

$$MSE_{ij} = \left(output_{reg_{ij}} - label_{reg_{ij}}\right)^2;$$

$$L\_r(W,b) = \frac{\sum_{k=1}^{m}(\sum_{i=0}^{n}\sum_{j=0}^{n}(MSE_{ij}*WM_{ij}))}{m}$$

10:   compute total loss:
$L_{all} = \alpha L\_c + \beta L\_r$
11:   update $W$;
12:   update $b$;
13:   $count += 1;$
14: **end while**

## EXPERIMENTS

Here, we evaluate (1) single-stream single classification (Single cls); (2) single-stream single regression (Single reg); (3) TS single classification (TwoStream cls); (4) TS single regression (TwoStream reg); and, (5) TS-MT models, all of which were implemented in Pytorch. All models employed identical training/test data. The initial learning rate was 0.001 and the batch size was 64. Training ended after 100,000 iterations (about 300 epochs).

TABLE 1
CONFUSION MATRIX UNDER TWO-CLASSFITION PROBLEM

| **Confusion Matrix** | | |
|---|---|---|
|  | **Predicted Positive** | **Predicted Negative** |
| **Real Positive** | True Positive (TP) | False Negative (FN) |
| **Real Negative** | False Positive (FP) | True Negative (TN) |

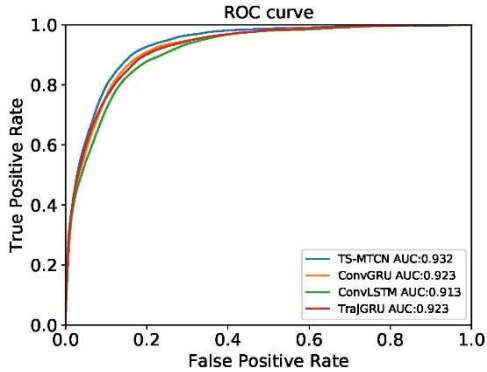

Fig. 5. The ROC curve

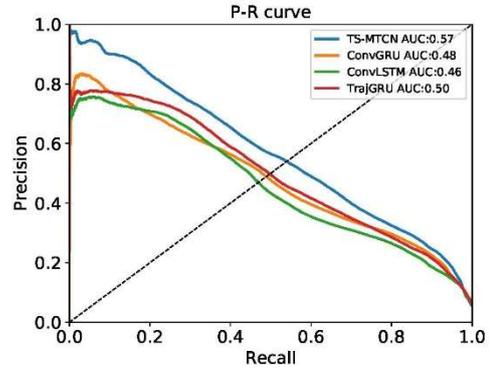

Fig. 6. The PR curve

### A. Datasets

The radar echo and satellite training and test datasets were collected in the Beijing-Tianjin-Hebei triangle in 2015. Because strongly convective weather did not occur on all days, and as we targeted convective storms, we generated a dataset including the data from 13 days of intensely convective weather. To distinguish the training, test, and verification data, information from 11, 1, and 1 day(s) was used for training, validation, and testing, respectively. We carry out this 4 times (4-fold cross validation, 4-CV for short) to compare the average and standard deviation of each evaluation criteria in Table 2, 3 and 4. Data were sliced using 5- and 3-frame sliding windows. After pixel-wise sampling, the sample size of the radar data was 1*5*54*54 and that of the satellite data was 13*3*27*27.

## B. Evaluation criteria

Several widely used methods were used to evaluate the models [the dBZ mean squared error (dBZ-MSE), probability of detection (POD), false alarm ratio (FAR), and critical success index (CSI)]. The dBZ-MSE is the average squared error of the difference between the predicted dBZ and the ground truth. CSI, POD, and FAR are similar to precision and recall (skill scores used in the machine learning field). The confusion matrix used to evaluate CSI, POD, and FAR is shown in Table I. We converted ground truths and predictions to a 1/0 matrix based on the output probabilities, and calculated hits (prediction = 1, ground truth = 1), misses (prediction = 0, ground truth = 1), and false alarms (prediction = 1, ground truth = 0); the CSI, POD, and FAR were respectively defined as follows:

$$POD = \frac{TP}{TP + FN}$$
$$FAR = \frac{FP}{TP + FP} \quad (9)$$
$$CSI = \frac{TP}{TP + FN + FP}$$

We compared forecast to actual labels and calculated CSI, POD, and FAR values. Reference [30] considered that the CSI was useful when dealing with low-frequency events such as severe weather, where the higher the CSI and the lower the FAR, the better the model.

TABLE II
COMPARSION OF TS-MTCN-C AND TS-MTCN-R NETWORKS WITH SINGLE_CLS, SINGLE_REG, TWOSTREAM_CLS AND TWOSTRAM_REG NETWORK ON THE SAMEDATASET

| Model | CSI | POD | FAR |
|---|---|---|---|
| Single_cls | 0.341 | 0.603 | 0.561 |
| Single_reg | 0.327 | 0.661 | 0.613 |
| TwoStream_cls | 0.351 | 0.713 | 0.591 |
| TwoStream_reg | 0.339 | 0.665 | 0.574 |
| **TS-MTCN-C** | **0.373** | **0.720** | **0.563** |
| **TS-MTCN-R** | **0.358** | **0.606** | **0.506** |

## C. Comparison among the new networks

For early warnings regarding strong convective storms, classification accuracy is somewhat more important than regression accuracy. To explore whether the TSMT method improved classification accuracy, we established Single_cls, Single_reg, TS_cls and TS_reg networks, and compared them using identical training, verification, and test sets (Table 2). The single-stream and TS networks were more accurate than the single-stream method, which was in turn better than the single-task method; satellite/radar data fusion and MT learning were valuable.

## D. Comparison of the new networks with other networks

We compared the classification and regression task performances of ConvLSTM, TrajGRU, ConvGRU, and TS-MTCN, and calculated CSI, POD, and FAR, as well as the number of parameters.

*1) Classification:* All results are shown in Table 3 and Figure 5. TS-MTCN classification (TS-MTCN-C) was the most accurate. Also, as the network is simple, the number of parameters was far lower versus other networks. Our network optimizes classification. First, the MT model incorporates cross-task correlations, which improves performance. Second, the TS network compensates for poor radar data coverage.

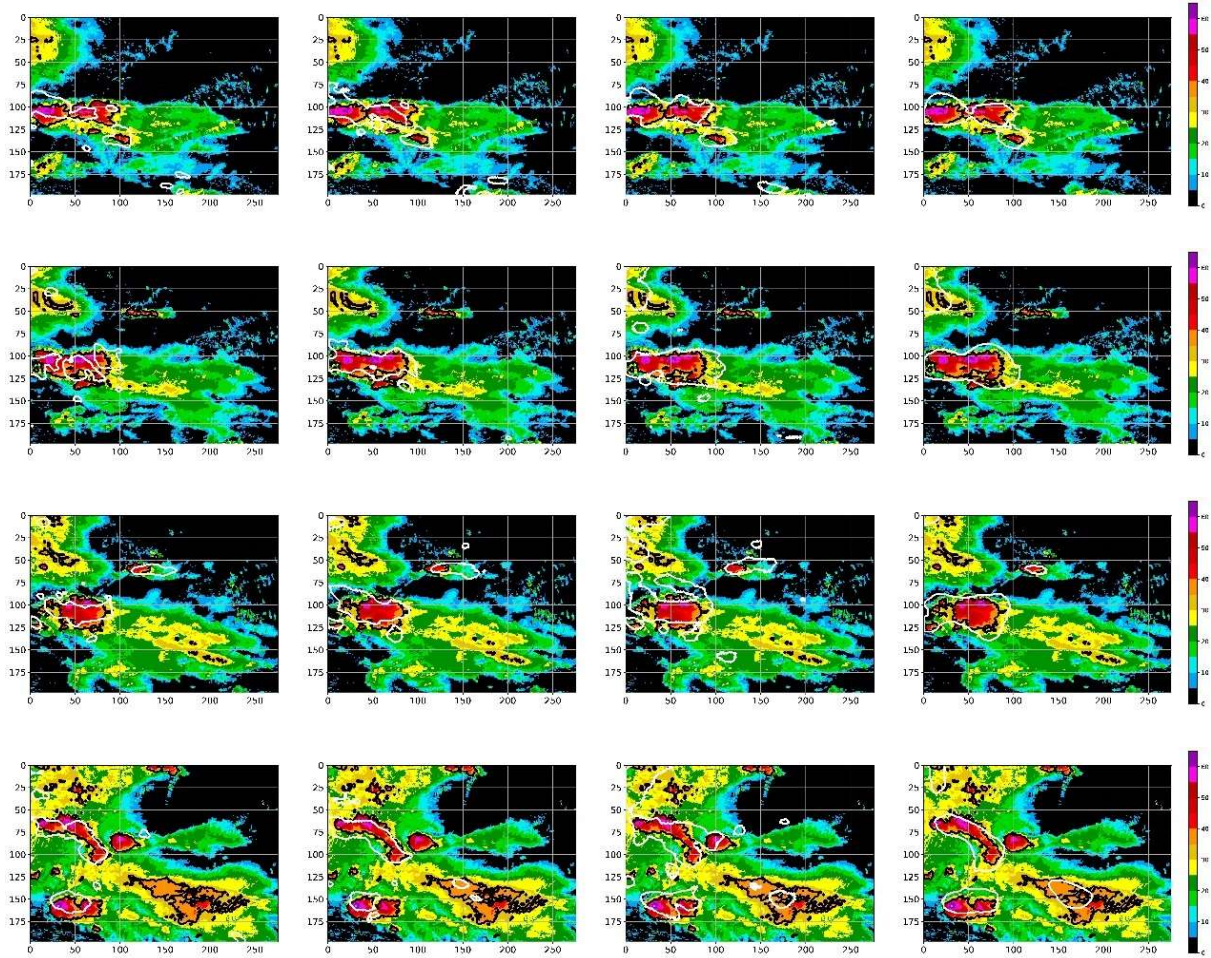

Fig. 4. From left to right, the classification nowcasting results of ConvLSTM, TrajGRU, ConvGRU and TS-MTCN-C. The white line areas in the figure indicates the area where the storm will be predicted in the next 30 minutes, and the black line areas indicates the ground truth

TABLE III

COMPARISON OF TWOSTRAM_REG NETWORK WITH CONVLSTM, TRAJGRU, AND CONVGRU NETWORK ON THE SAME DATASET

| Model | CSI | POD | FAR | MSE | Number of parameters |
|---|---|---|---|---|---|
| ConvLSTM | 0.329 | 0.492 | 0.503 | 85.246 | 18381205 |
| TrajGRU | 0.343 | 0.514 | 0.492 | 83.126 | 10143701 |
| ConvGRU | 0.329 | 0.716 | 0.621 | 104.911 | 14557625 |
| **TS-MTCN-C** | **0.373** | **0.720** | **0.563** | **-** | **1309051** |
| **TwoStream_reg** | **0.324** | **0.635** | **0.602** | **89.788** | **1205649** |

When combined with satellite data, useful characteristics can be extracted. Finally, the TS MT CNN is simple in structure and easy to train. In particular, pixel-wise sampling solves the problem of the small sample size; the network is welltrained using a small number of data points and parameters. Figure 5 shows network receiver operator characteristic (ROC) and precision-recall (PR) curves for one of the 4-CV (others are similar). As ConvLSTM, TrajGRU, and ConvGRU do not yield classification probabilities, we normalized the radar echo values to 0-1 when plotting the curves. Although our network performs somewhat better than the others in terms of ROCs, the PR space reveals the clear advantage of our network. The upper-left-hand corner is the optimal ROC space, wherein all curves are near-optimal. The upper-right-hand corner is the optimal PR space; although our network has achieved the best results, the convective nowcasting problem still needs further study, since the areas under the PR curve of all algorithms are less than 0.6.

Figure 4 compares the qualitative classifications derived using ConvLSTM, TrajGRU, ConvGRU, and TS-MTCN-C at 09:00, 09:30, 10:00, and 10:30 on July 29. In the four prediction columns, the black lines are the ground truths and the white lines are the forecast areas. When Figure 4 and Table 3 are viewed together, it is clear that the ConvLSTM, TrajGRU, and ConvGRU networks generate more false-negatives than ours. Especially at 10:00, our network accurately predicted a small region of strongly convective weather; ConvLSTM, TrajGRU, and ConvGRU were less accurate at small scales and tended to overstate the affected areas.

*2) Regression:* Figure 7 and Table 3 compare qualitative regressions of ConvLSTM, TrajGRU, ConvGRU, and our network. The first column corresponds to the regression image of ConvLSTM, the second and third columns correspond to the regression images of TrajGRU and ConvGRU, the fourth column corresponds to the image of our model, and the last final column corresponds to the ground truth. Our results are close to those of other networks, but our network required far fewer parameters

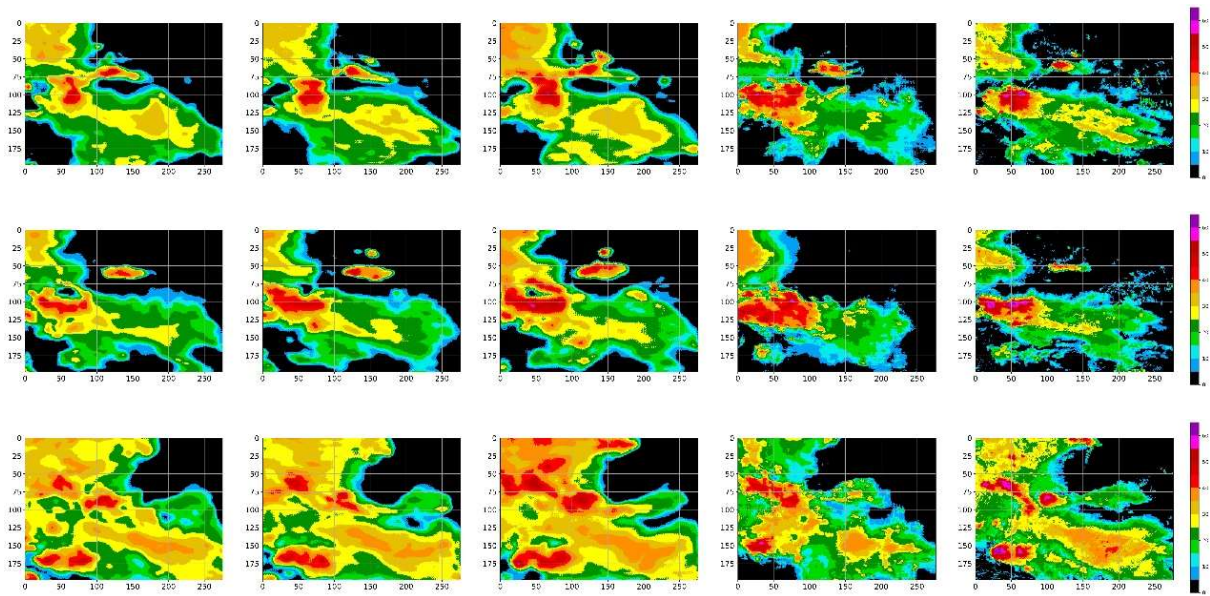

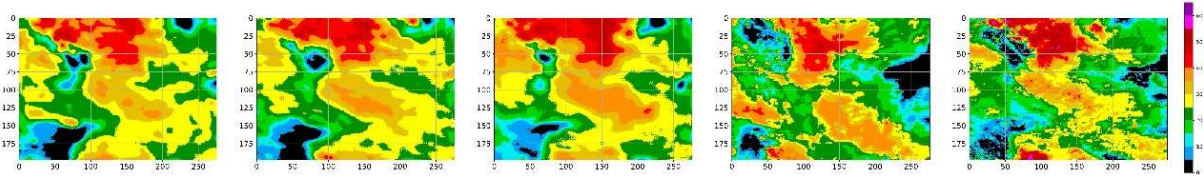

Fig. 7. From left to right, the regression nowcasting results of ConvLSTM, TrajGRU, ConvGRU, TwoStream_reg and Ground truth

CONCLUSIONS

We used radar and satellite data to nowcast severe convective weather. We employed a TSMT CNN to not only consider the relationships between regression and classification tasks, but also a diverse range of data sources. This significantly improved the predictive accuracy of the classification. However, prediction of convective storms remains challenging and is urgently required; to achieve this, our model requires improvement. Although regression is useful for auxiliary classification, the predictive accuracy must be improved. In the future, we will use a simplified, dense, optical-flow-based frame rate conversion algorithm to generate synthetic satellite data at 6-min intervals, and then build an operational nowcasting system. Following this, we will set multi-category rather than two-category tasks.